\documentclass{article} 
\usepackage{url}
\usepackage{algorithm}
\usepackage{algorithmic}
\usepackage{amssymb}
\usepackage{amsmath}
\usepackage{amsfonts}
\usepackage{bm}
\usepackage{booktabs}
\usepackage{subfigure}
\usepackage{graphicx}
\usepackage{amsthm}
\usepackage{etoolbox}
\usepackage{epstopdf}
\patchcmd{\thebibliography}{\section*{\refname}}{}{}{}

\title{Power-Law Graph Cuts}

\author{
Xiangyang Zhou, Jiaxin Zhang, Brian Kulis \\
Department of Computer Science and Engineering\\
The Ohio State University\\
\texttt \{zhou.1226, zhang.2725\}@osu.edu, kulis@cse.ohio-state.edu
}

\begin{document}

\maketitle

\begin{abstract}
Algorithms based on spectral graph cut objectives such as normalized cuts, ratio cuts and ratio association have become popular in recent years because they are widely applicable and simple to implement via standard eigenvector computations. Despite strong performance for a number of clustering tasks, spectral graph cut algorithms still suffer from several limitations: first, they require the number of clusters to be known in advance, but this information is often unknown \textit{a priori}; second, they tend to produce clusters with uniform sizes.  In some cases, the true clusters exhibit a known size distribution; in image segmentation, for instance, human-segmented images tend to yield segment sizes that follow a power-law distribution. In this paper, we propose a general framework of power-law graph cut algorithms that produce clusters whose sizes are power-law distributed, and also does not fix the number of clusters upfront.  To achieve our goals, we treat the Pitman-Yor exchangeable partition probability function (EPPF) as a regularizer to graph cut objectives. Because the resulting objectives cannot be solved by relaxing via eigenvectors, we derive a simple iterative algorithm to locally optimize the objectives.  Moreover, we show that our proposed algorithm can be viewed as performing MAP inference on a particular Pitman-Yor mixture model. Our experiments on various data sets show the effectiveness of our algorithms.

\end{abstract}

\section{Introduction}

As one of the most fundamental problems in machine learning, clustering has received a considerable amount of attention and has applications in data mining, computer vision, statistics, social sciences, and others.
Spectral graph cut algorithms such as normalized cuts \cite{shi2000normalized}, ratio cut~\cite{chan1994spectral} and ratio association~\cite{shi2000normalized, dhillon2007weighted} are one of the most studied and utilized classes of clustering methods.
These algorithms aim to cluster data by first constructing a similarity graph based on the given data, then ``cutting" the graph into groups of nodes according to a graph-theoretic objective.
Normalized cuts has been widely used in the computer vision community for image segmentation~\cite{shi2000normalized} and other problems~\cite{greig1989exact} while ratio cut has been applied in circuit layout~\cite{chan1994spectral}.
Though these graph cut problems can be shown to be NP-hard, several effective algorithms have been proposed, including eigenvector-based approaches~\cite{shi2000normalized} as well as methods based on kernel k-means~\cite{dhillon2007weighted}.

Despite the success of spectral graph cut algorithms, they do suffer from several important limitations.
For one, they require the number of clusters to be known before running the algorithm, but in many applications the number of clusters is not known \textit{a priori}.
More importantly, many graph cut objectives, such as the normalized cut objective and the ratio cut objective, favor clusters of equal size or degree, which typically leads these algorithms to produce clusters with nearly uniform sizes.
Consider image segmentation, the canonical application of normalized cuts.  As shown in~\cite{sudderth2008shared}, human-segmented images yield segments that are far from uniform; in fact, they follow a \textit{power-law distribution} in terms of their segment sizes.
Power-law distributions arise frequently in a number of other clustering applications as well.
For instance, because income follows a power-law distribution, attempting to cluster individuals into income brackets using census data would likely fail when applying standard clustering techniques.
Other phenomena exhibiting power-law distributions include the populations of cities, the intensities of earthquakes, and the sizes of power outages \cite{clauset2009power}.
These applications---and the lack of existing graph clustering methods that specifically encourage power-law cluster size structure---motivate our work.

In this paper, we propose a general framework of power-law graph cut algorithms that encourages cluster sizes to be power-law distributed, and does not fix the number of clusters upfront.
To achieve both goals, we borrow ideas from Bayesian nonparametrics~\cite{hjort}, which provide a principled way to automatically infer both the parameters of a model as well as its complexity.
We observe that the Pitman-Yor process~\cite{pitman1997two}, a Bayesian nonparametric prior that generalizes the Chinese restaurant process,  yields clusters whose sizes follow a power-law distribution.
We treat the Pitman-Yor exchangeable partition probability function (EPPF)~\cite{pitman} as a \textit{regularizer} for graph cut objectives, so that the resulting objectives favor clusters that both have a small graph cut objective value as well as a power-law cluster size structure.

Algorithmically, incorporating the Pitman-Yor EPPF into existing cut formulations results in an optimization problem where standard spectral methods are no longer applicable.
Inspired by the connection between spectral graph cut objectives and weighted kernel k-means~\cite{dhillon2007weighted}, we derive a simple k-means-like iterative algorithm to optimize several power-law graph cut objectives.
As with k-means, our proposed algorithm is guaranteed to converge to a local optima in a finite number of steps.
We further demonstrate that our graph cut problem may be viewed precisely as a MAP problem for a particular Pitman-Yor Gaussian mixture model.
Finally, to demonstrate the utility of our algorithm, we perform extensive experiments using power-law normalized cuts on synthetic datasets, real-world data with power-law structure, and image segmentation.


\textbf{Related Work:}
Small-variance asymptotics have recently been extended to Bayesian nonparametric models to yield simple k-means-like algorithms~\cite{kulis2012revisiting,broderick2012mad}; one of the applications of that line of work is a normalized cut algorithm that does not fix the number of clusters upfront~\cite{kulis2012revisiting}.
However, that approach cannot be directly applied to Pitman-Yor process mixture models, as small-variance asymptotics on the Pitman-Yor process model fail to capture any power-law characteristics.

The most related work to ours is~\cite{DBLP:journals/corr/FanZC13}, an algorithm for scalable power-law clustering based on adapting k-means.
Specifically, the authors propose a power-law data clustering algorithm based on modifying the Pitman-Yor process and performing a small-variance asymptotic analysis on the modified Piman-Yor process.
However, their objective function does not guarantee the generation of a power-law distributed cluster sizes and the optimal clustering solutions for their objective are often trivial.
We will discuss this method further in Section \ref{sec:comparison_theoretical} and Section \ref{Sec:exp}.

Finally, the work of~\cite{sudderth2008shared} introduces a model for segmentation based on Pitman-Yor priors, but it is specific to the image domain whereas our method is a general graph clustering algorithm.


\section{Background}

We begin with a brief discussion about spectral graph cut algorithms and their connection to weighted kernel k-means.


\subsection{Spectral graph cut algorithms}

In the graph clustering setting, we are given an undirected weighted graph $G = (\mathcal{V}, \mathcal{E})$, in which $\mathcal{V} = \{v_1, ..., v_n\}$ denotes vertices and $\mathcal{E}$ denotes edges.
The weight of an edge between two vertices represents their similarity.
The corresponding adjacency matrix $A$ is a $|\mathcal{V}|$-by-$|\mathcal{V}|$ matrix whose entry $A_{ij}$ represents the weight of the edge between $v_i$ and $v_j$.

The idea behind graph cuts is to partition the graph into $k$ disjoint clusters such that the edges within a cluster have high weight and the edges between clusters have low weight.
Several different graph cut objectives have been proposed~\cite{shi2000normalized,chan1994spectral,kernighan1970efficient}, among which normalized cuts~\cite{shi2000normalized} and ratio cut~\cite{chan1994spectral} are two of the most popular.
Denote
\[	\text{cut}( \mathcal{V}_1, \mathcal{V}_2 ) = \sum_{i \in \mathcal{V}_1, j \in\mathcal{V}_2} A_{ij},		\]
i.e., the sum of the edge weights between $\mathcal{V}_1$ and $\mathcal{V}_2$, and
\[	\text{deg}( \mathcal{V}_1) = \text{cut}( \mathcal{V}_1,\mathcal{V} ),		\]
the sum of all edge weights between $\mathcal{V}_1$ and $\mathcal{V}$.
Normalized cuts (sometimes called $k$-way normalized cuts) aims to minimize the cut relative to the degree of the cluster. The objective can be expressed as
\[	\text{NCut}(G) = \min_{\mathcal{V}_1, ...,\mathcal{V}_k}	\sum_{i=1}^k \frac{\text{cut}(\mathcal{V}_i, \mathcal{V}/\mathcal{V}_i)}{\text{deg}(\mathcal{V}_i)}.	\]
While this objective can be shown to be NP-complete, a relaxation of it can be globally optimized using spectral methods by computing the first $k$ eigenvalues of the normalized Laplacian constructed from the adjacency matrix $A$~\cite{yu2003multiclass}.

The ratio cut objective differs from normalized cuts in that it seeks to minimize the cut between clusters and the remaining vertices. It is expressed as
\[	\text{RCut}(G) = \min_{\mathcal{V}_1, ...,\mathcal{V}_k}	\sum_{i=1}^k \frac{\text{cut}(\mathcal{V}_i, \mathcal{V}/\mathcal{V}_i)}{|\mathcal{V}_i|}.	\]
Note that there are also other graph partitioning objectives that fall under this framework (see, e.g., Section 3 of~\cite{dhillon2007weighted} which generalizes association and cut problems to weighted variants), and our approach can also be applied to these objectives.


\subsection{Weighted kernel k-means and graph cuts}			\label{sec:connection}

Consider the k-means objective function with clusters $\ell_1, \dots, \ell_k$:
$$		\sum_{c=1}^k \sum_{ \boldsymbol{x} \in l_c } 	\| 	\boldsymbol{x} - \boldsymbol{\mu}_c	\|^2,		$$
where $\boldsymbol{\mu}_c =  (1/|\ell_c|)	\cdot \sum_{ \boldsymbol{x}\in \ell_c } \boldsymbol{x}$.
It is straightforward to extend this to the weighted setting by introducing a weight $w_i$ for each data point, which yields the following:
$$		\sum_{c=1}^k \sum_{\boldsymbol{x}\in \ell_c} w_i \| \boldsymbol{x} -  \boldsymbol{\mu}_c \|^2,		$$
where now the mean $\boldsymbol{\mu}_c$ is the weighted mean $\boldsymbol{\mu}_c = \sum_{\boldsymbol{x}_i\in \ell_c}w_i \boldsymbol{x}_i / \sum_{\boldsymbol{x}_i\in \ell_c}w_i $.
Further, we can replace the original data with mapped data $\phi(\boldsymbol{x})$ and treat the entire problem in kernel space by expressing both the k-means algorithm, along with the objective, in terms of inner products.
This is necessary for the connection to graph cuts.

Dhillon et al. \cite{dhillon2007weighted} showed that there is a connection between the weighted kernel k-means objective and several spectral graph cut objectives. We will discuss in particular the connection to normalized cuts. Define the degree matrix $D$ as the diagonal matrix whose entries $D_{ii}$ are equal to the degree of node $i$.
The surprising fact established in~\cite{dhillon2007weighted} is that normalized cuts and weighted kernel k-means are mathematically equivalent, in the following sense:
if $A$ is an adjacency matrix, then the normalized cuts objective on $A$ is equivalent to the weighted kernel k-means objective (plus a constant) on the kernel matrix $K= \rho D^{-1} + D^{-1}AD^{-1}$, where $\rho$ is chosen such that $K$ is a positive semi-definite matrix, and where the weights of the data points are equal to the degrees of the nodes.
Thus, for the purposes of minimizing the weighted (kernel) k-means objective function, we can effectively interchange the objective with the normalized cut objective, i.e.,
\begin{equation}\label{eq:graphcutkmeansequiv}
		\min \sum_{c=1}^k \sum_{  \boldsymbol{x} \in \ell_c  } w_i \| \boldsymbol{x} -  \boldsymbol{\mu}_c \|^2 			 \equiv
		\min \sum_{c=1}^k \frac{  \text{cut}(  \mathcal{V}_c,  \mathcal{V} \slash \mathcal{V}_c )  }{  \text{deg}(\mathcal{V}_c)  }
\end{equation}
for the appropriate definition of the kernel matrix. In particular, this result gives an algorithm for monotonically minimizing the normalized cut objective---we just form the appropriate kernel and set the weights to the degrees, and then run weighted kernel k-means on that kernel matrix. Similar equivalences hold for both the ratio cut and ratio association objectives---by forming appropriate kernels and weights, the graph cut objectives can be shown to be mathematically equivalent to the weighted kernel k-means objective.

\section{The Power-law Normalized Cut Objective}

Our goal is to propose and study a new set of graph cut objectives that produce power-law distributed cluster sizes. In order to achieve this, we will borrow some key ideas from Bayesian nonparametrics.  More specifically, we look at the Pitman-Yor process~\cite{pitman1997two}, a generalization of the Chinese Restaurant Process that specifically yields power-law distributed cluster sizes. For simplicity, we will focus on the normalized cut objective as an example. One can simply replace the normalized cut objective with other graph cut objectives to obtain other power-law graph cut algorithms in our framework.


\subsection{Pitman-Yor EPPF}

The canonical Bayesian nonparametric clustering prior is the Chinese restaurant process (CRP)~\cite{hjort}.
It yields a distribution on clusterings such that the number of clusters are not fixed, and where the sizes of the clusters decay exponentially.
The description of the CRP is as follows: customers enter a restaurant with an infinite number of tables (each table corresponds to a cluster).
The first customer sits at the first table.
Subsequent customers sit at tables with probability proportional to the number of seated customers at that table, and with probability proportional to $\alpha$ sit at a new table.
The Pitman-Yor process leads to an extension of the CRP such that the cluster sizes instead follow a power-law distribution.
In this modified version of the CRP, when customers sit down at tables, they sit at an existing table with probability proportional to the number of existing occupants minus $\theta$ ($0 \leq \theta < 1$), and at a new table with probability proportional to $k \cdot \theta + \alpha$, where $k$ is the current number of occupied tables.
Thus, as the number of tables $k$ increases, there is a higher probability of starting a new table; this leads to the heavier power-law distribution of cluster sizes.

One can explicitly write down the probability of observing a particular seating arrangement under the Pitman-Yor CRP, and the resulting formula is known as the Pitman-Yor exchangeable partition probability function (EPPF)~\cite{pitman}.
If we let $Z$ be an indicator matrix for the resulting clustering, then the probability distribution of $Z$ under the Pitman-Yor CRP is expressed by the following unintuitive and somewhat cumbersome form:
\begin{equation}
	\label{eq:PitmanYorEPPF}
		p(Z | \alpha, \theta) = \frac{ [\alpha + \theta]_{k-1, \theta} }{[\alpha + 1]_{n-1}} \cdot \prod_{c=1}^k [ 1 - \theta ]_{n_c - 1}
	\end{equation}
where
\begin{equation*}
	[x]_{m,a} = \begin{cases}
				1\qquad	 						 &m = 0 			
				\\x \cdot (x + a) \cdots ( x + (m - 1)a ) 	\, &m = 1,2, ... ,
				\end{cases}
	\end{equation*}
$n_c$ is the size of cluster $c$, and $[x]_m$ is defined as $[x]_{m,1}$.
One can verify that, when $\theta = 0$, we actually obtain the original CRP probability distribution.
One can also show that the expected number of clusters under this distribution is $\mathcal{O}(\alpha n^\theta)$, and that we obtain the desired power-law cluster size distribution.


\subsection{Power-law normalized cut objective}\label{sec:ObjectiveFunction}

To obtain power-law distributed cluster sizes within a graph clustering setting, we treat the Pitman-Yor EPPF as a regularizer for the cluster indicator matrix of normalized cuts.
Then our resulting objective is given as below:
\begin{align}		\label{eq:powerlawobjective}
			\min_{Z, k} &\sum_{c=1}^k \frac{\text{cut} (\mathcal{V}_c, \mathcal{V}\backslash \mathcal{V}_c)}{\text{deg}(\mathcal{V}_c )}
										+ \lambda \cdot r_{\alpha, \theta}(Z),										
		\\	& r_{\alpha, \theta}(Z) = -\ln p(Z | \alpha, \theta)			\nonumber					
	\end{align}
where $Z = [z_1,...,z_n]$ is the indicator for the cluster assignment of each node, $ r_{\alpha, \theta}(Z)$ is the negative log of the Piman-Yor EPPF and $\lambda$ is a tradeoff between the original graph cut objective and the regularization term.
The first term is the standard normalized cut objective.
The desired power-law distributed partition would give a high value of the Pitman-Yor EPPF and thus a low value of the second term.
Therefore, the clustering result that minimizes this objective should give a partition of the graph such that both similarity information is preserved and cluster sizes are power-law distributed.


\section{Optimization}

The objective function~\eqref{eq:powerlawobjective} defined in the previous section enforces a tradeoff between standard normalized cuts and a preference for power-law cluster size structure.
We now turn to optimization of the resulting objective.


\subsection{The vector case}

The first observation that we can make is that spectral methods will not apply to our proposed objective.
Recall that for the normalized cut objective, a standard approach is to relax the cluster indicator matrix $Z$ to be continuous, leading to a simple eigenvector problem that can be optimized globally.
When applying such a technique to the power-law normalized cut objective, one would need to incorporate the regularization term appropriately into the trace maximization problem that emerges from the spectral solution, but this turns out to be impossible.

Instead we must turn to the other main optimization strategy for normalized cuts---namely the equivalence to weighted kernel k-means---and we will adapt the weighted kernel k-means algorithm for our problem.
To start, in this section we will derive a k-means-like algorithm for the following regularized k-means problem:
\begin{displaymath}
	\min_{Z,k} \sum_{c=1}^k \sum_{\bm{x}_i \in \ell_c} w_i \|\bm{x}_i - \bm{\mu}_c\|^2  + \lambda \cdot r_{\alpha, \theta}(Z),
	\end{displaymath}
where the means $\bm{\mu}_c$ are the weighted means of the points in $\ell_c$ as in standard weighted k-means as discussed in Section~\ref{sec:connection}.
Once we have obtained the algorithm for this case, we can easily extend the connection between normalized cuts and weighted kernel k-means to obtain an algorithm for monotonic local convergence of the power-law normalized cut objective.
Note that this treatment is equally applicable to the ratio cut and ratio association objectives.

We observe that, when the cluster indicators $Z$ are fixed, the weighted mean is justified in the above objective since it is the best cluster representative for each cluster in terms of the objective function, i.e., for fixed $Z$ and any choice of $c$, the regularizer is constant and we have by simple differentiation
\begin{displaymath}
	\sum_{\bm{x}_i \in \ell_c} w_i \|\bm{x}_i - \bm{\mu}_c\|^2 = \min_{\bm{m}} \sum_{\bm{x}_i \in \ell_c} w_i \|\bm{x}_i - \bm{m}\|^2.
	\end{displaymath}
Therefore, the updates on $\bm{\mu}_c$ will be exactly as in standard weighted k-means.

The other step is the update on the indicators $Z$.
In standard k-means, these updates are derived by fixing the means and minimizing the k-means objective function with respect to each $z_i$, which yields the usual k-means assignment step.
The Pitman-Yor EPPF regularizer makes the assignment updates somewhat less trivial, but it is still fairly straightforward.
For each data point we consider the objective function when assigning that point to every existing cluster, as well as to a new cluster, and assign to the cluster that results in the smallest objective function.
The regularizer effectively adds a ``correction" to each distance computation $w_i \|\bm{x}_i - \bm{\mu}_c\|^2$.
Let $n_c$ be the number of points in cluster $c$.
After going through the algebra, we arrive at the following: if $\boldsymbol{x}_i$ is currently assigned to $\ell_c$, then we have that the distance to another cluster $\ell_{c^\prime}$ (ignoring constants, which do not affect re-assignment) is given by the following cases:
\begin{itemize}
	\itemsep0em
		\item If $c^\prime = c$:
			\[	d(\boldsymbol{x}_i, c^\prime) =	w_i  \| \boldsymbol{x}_i - \boldsymbol{\mu}_{c^\prime} \|^2.     \]
		\item if $n_c > 1$ and $c^\prime$ is an existing cluster,
			\[		d(\boldsymbol{x}_i, c^\prime)
				= w_i  \| \boldsymbol{x}_i - \boldsymbol{\mu}_{c^\prime} \|^2 + \lambda\cdot\ln \left( \frac{n_c-1-\theta}{n_{c^\prime} - \theta}\right).	 \]
		\item if $n_c = 1$ and $c^\prime$ is an existing cluster,
			\[	d(\boldsymbol{x}_i, c^\prime) =
				w_i  \| \boldsymbol{x}_i - \boldsymbol{\mu}_{c^\prime} \|^2 + \lambda\cdot\ln \left( \frac{\alpha+ (k-1)\theta}{n_{c^\prime} - \theta}	\right).	\]	
		\item if $n_c > 1$ and $c^\prime$ is a new cluster
			\[	d(\boldsymbol{x}_i, c^\prime) =\lambda\cdot\ln\left(\frac{n_c-1-\theta}{\alpha+ k\theta}\right) . 	 \]
		\item  if $n_c = 1$ and $c^\prime$ is a new cluster
			\[	d(\boldsymbol{x}_i, c^\prime) =\infty.	\]	
	\end{itemize}
Observe that the distance to new clusters goes down as $k$ increases, which is analogous to the property in the Pitman-Yor version of the Chinese restaurant process of being more likely to start a new table as the number of tables increases.
In a similar way, when computing the distance to existing clusters, the distance becomes smaller as the cluster gets larger (i.e., as $n_{c^\prime}$ goes up), leading to the ``rich gets richer'' behavior.
Finally, whenever a new cluster is started by some point $\bm{x}_i$, we immediately set the mean to be $\bm{x}_i$.
See Algorithm~\ref{algorithm} for a full specification.
Note that, analogous to the convergence proof of k-means, one can easily show that this algorithm monotonically decreases the regularized k-means objective until local convergence.


\subsection{Power-law normalized cut algorithm}\label{sec:powerlawncutalgorithm}

Recall that in Section \ref{sec:connection} we discussed the equivalence between the graph cuts formulation and the weighted kernel k-means objective, as in (\ref{eq:graphcutkmeansequiv}).
With this equivalence in hand, the extension from the vector case to the power-law graph cut objectives follows easily: we simply replace the weighted k-means term with a graph cuts term, which gives exactly the same objective with our power-law graph cuts objective in (\ref{eq:powerlawobjective}) up to a constant; then we apply Algorithm \ref{algorithm} in kernel space to solve the resulting optimization problem.

More specifically, given a graph $G = (\mathcal{V}, \mathcal{E})$ with adjacency matrix $A$, our power-law normalized cut algorithm is described as follows:
\begin{itemize}
	\item Compute the degree matrix $D$ from $A$ as the diagonal matrix whose entries $D_{ii}$ are equal to the degree of node $i$.
	\item Compute the kernel matrix $K$ from $A$ using $K = \rho D^{-1} + D^{-1}AD^{-1}$.
	\item Run Algorithm \ref{algorithm} in kernel space with kernel $K$ and weights given by the degrees of the nodes.
			The power-law normalized cut clustering result is then obtained directly from Algorithm \ref{algorithm}.
\end{itemize}

In kernel space, the regularized distance remains unchanged.
The only change is that now we need to compute $\| \phi(\boldsymbol{x}_i) - \boldsymbol{\mu}_c \|^2$ instead of  $\| \boldsymbol{x}_i - \boldsymbol{\mu}_c \|^2$.
We expand the last distance computation and use the formula for $\boldsymbol{\mu}_c$ and obtain:
\begin{align*}
	 \phi(\boldsymbol{x}_i)\cdot \phi(\boldsymbol{x}_i) &- \frac{2\sum_{\boldsymbol{x}_j\in l_c}w_j \phi(\boldsymbol{x}_i)\cdot \phi(\boldsymbol{x}_j) }{\sum_{\boldsymbol{x}_j\in l_c}w_j}
	 			+  	\frac{\sum_{\boldsymbol{x}_j \boldsymbol{x}_k \in l_c}w_jw_k \phi(\boldsymbol{x}_j)\cdot \phi(\boldsymbol{x}_k) }{(\sum_{\boldsymbol{x}_j\in l_c}w_j)^2}.
\end{align*}

Using the kernel matrix $K$, the above may be written as:
\begin{align*}
K_{ii} - \frac{2\sum_{\boldsymbol{x}_j\in l_c}w_jK_{ij} }{\sum_{\boldsymbol{x}_j\in l_c}w_j}  +  \frac{\sum_{\boldsymbol{x}_j \boldsymbol{x}_k \in l_c}w_jw_k K_{jk} }{(\sum_{\boldsymbol{x}_j\in l_c}w_j)^2}.
\end{align*}
We note that, as when applying weighted kernel k-means to the standard normalized cut problem~\cite{dhillon2007weighted}, each iteration of Algorithm~\ref{algorithm} when applied in kernel space with $K$ requires time $O(|{\mathcal E}|)$, making it very scalable for applications to large graphs.
Also note that by using an appropriate kernel matrix $K$, we can utilize other graph cut objectives in this framework.

\begin{algorithm}[!h]
\caption{Power-law-means (vector case)}
\begin{algorithmic}[1]\label{algorithm}
\renewcommand{\algorithmicrequire}{\textbf{Input:}}
\renewcommand{\algorithmicensure}{\textbf{Output:}}
\REQUIRE {$\boldsymbol{x_{1}}, ..., \boldsymbol{x_{n}}$: data points;\; $w_1, ..., w_n$: weights; \;$\lambda$: trade-off parameter;\; $\alpha, \theta$: Pitman-Yor EPPF parameters}
\ENSURE {Clustering $\ell_1, ..., \ell_k$;\; $k:$ number of clusters}
\STATE Init. $k=1, \ell_1 = \{ \boldsymbol{x_{1}}, ..., \boldsymbol{x_{n}}\}, \boldsymbol{\mu_1}$ the global mean
\STATE Init. cluster indicators $z_i = 1$ for all $i=1, ...,n$.
\STATE Repeat \ref{repeatBegin1} to \ref{repeatEnd1} until convergence.
\FOR {each data point $\boldsymbol{x_i}$, suppose $\boldsymbol{x_i}$ is currently assigned to cluster $\ell_c$ } \label{repeatBegin1}
\IF {$n_c = 1$, i.e. $\boldsymbol{x}_i$ is a singleton cluster}
\STATE compute its ``regularized'' distance $d(\boldsymbol{x}_i, c^\prime)$ to the other clusters according to the following:
	\begin{itemize}
		\item If  $c^\prime = c$, $d(\boldsymbol{x}_i, c^\prime) = 0$.
		\item  if $c^\prime \not= c$ and $\ell_{c^\prime}$ is an existing cluster,					
\\				$d(\boldsymbol{x}_i, c^\prime)
					=w_i  \| \boldsymbol{x}_i - \boldsymbol{\mu}_{c^\prime} \|^2 + \lambda\cdot\ln ( \frac{\alpha+ (k-1)\theta}{n_{c^\prime} - \theta}	)$		
		\item  if $\ell_{c^\prime}$ is a new cluster, $d(\boldsymbol{x}_i, c^\prime) =\infty$
	\end{itemize}
\ELSE
\STATE compute its ``regularized" distance $d(\boldsymbol{x}_i, c^\prime)$ to the other clusters according to the following:
	\begin{itemize}
		\item If  $c^\prime = c$, $d(\boldsymbol{x}_i, c^\prime) = w_i  \| \boldsymbol{x}_i - \boldsymbol{\mu}_c^\prime \|^2 $.
		\item  if $c^\prime \not= c$ and $\ell_{c^\prime}$ is an existing cluster,						
\\				$d(\boldsymbol{x}_i, c^\prime)
					=w_i  \| \boldsymbol{x}_i - \boldsymbol{\mu}_{c^\prime} \|^2 + \lambda\cdot\ln (\frac{n_c-1-\theta}{n_{c^\prime} - \theta}	)$		
		\item  if $\ell_{c^\prime}$ is a new cluster,
				\\$d(\boldsymbol{x}_i, c^\prime) =  \lambda \cdot \ln (   \frac{ n_c-1-\theta }{ \alpha+ k\theta }    )$
	\end{itemize}
\ENDIF
\STATE Assign $\boldsymbol{x}_i$ to the cluster corresponding to the smallest regularized distance. Update $Z$:
	\begin{equation*}	\label{eq:updateIndicator}
			z_i = \arg \min_{c^\prime} d(\boldsymbol{x}_i, c^\prime).
	\end{equation*}
\IF{$z_i$ corresponds to a new cluster}
\STATE set $k \leftarrow  k+1, z_i =k$, and $\boldsymbol{\mu_k} = \boldsymbol{x}_i$.
\ENDIF
\ENDFOR
\FOR{each cluster $\ell_c$}
\STATE Update $\boldsymbol{\mu}_c$ based on the weighted mean of the data points in cluster $\ell_c$:
	\begin{equation*}	\label{eq:updateClusterMean}
		\boldsymbol{\mu}_c = \frac{\sum_{\boldsymbol{x}\in \ell_c}{w_i \boldsymbol{x}_i}}{\sum_{\boldsymbol{x}\in \ell_c} w_i }.
	\end{equation*}
\ENDFOR	\label{repeatEnd1}
\end{algorithmic}
\end{algorithm}





\subsection{Connection to Pitman-Yor MAP inference}

Finally, we briefly consider the connections between our proposed objective and a simple Pitman-Yor process mixture model.
Consider the following Bayesian nonparametric generative model:
\begin{align*}
			   Z  &\sim \text{PYCRP}(\alpha,\theta); \quad \quad
	\\	\bm{x}_i  &\sim \mathcal{N}(\bm{\mu}_{z_i},(\sigma/ (2 w_i)) I), \quad i = 1, ..., n,
\end{align*}
where PYCRP refers to the Pitman-Yor Chinese Resturant Process.
To perform MAP inference, we can write down the  joint likelihood and maximize with respect to the relevant parameters:
\begin{align*}
	&\text{argmax}_{Z,k,\bm{\mu}} p(X,Z) \\
	& \equiv \text{argmin}_{Z,k,\bm{\mu}} - \ln  \bigg(   \prod_{i=1}^n {\mathcal N}(\bm{\mu}_{z_i},\frac{\sigma}{2 w_i} I)  \cdot
	p(Z\,|\,\alpha,\theta)  \bigg)\\
	& \equiv \text{argmin}_{Z,k,\bm{\mu}} \frac{1}{\sigma} \sum_{c=1}^k \sum_{\bm{x}_i \in \ell_c} w_i \|\bm{x}_i - \bm{\mu}_c\|^2 - \ln p(Z\,|\,\alpha,\theta)\\
	& \equiv \text{argmin}_{Z,k,\bm{\mu}} \sum_{c=1}^k \sum_{\bm{x} \in \ell_c} w_i \|\bm{x}_i - \bm{\mu}_c\|^2 + \lambda \cdot r_{\alpha,\theta}(Z),
\end{align*}
where $\lambda = \sigma$.
Note that the minimization with respect to $\bm{\mu}$ yields precisely the weighted means, and so based on the equivalence between weighted kernel k-means and normalized cuts, we can see that our proposed objective function may be viewed in a MAP inference framework.
This framework also justifies the use of the log of the Pitman-Yor EPPF as a regularizer.


\subsection{Comparison to existing power-law clustering algorithm $pyp$-means}\label{sec:comparison_theoretical}

In \cite{DBLP:journals/corr/FanZC13}, the authors propose a different objective for power-law data clustering, namely:
\begin{align*}
	\arg \min_{l_1, ..., l_n} \sum_{c=1}^k \sum_{i \in l_c} \| \bm{x}_i - \bm{\mu}_c\| + (\lambda - \ln k \cdot \theta) k,
\end{align*}
which adds a $-k\ln k \cdot \theta$ term to the \textit{dp-means} objective function \cite{kulis2012revisiting}.
While this objective function does incorporate the number of clusters into the optimization, it does not require or encourage the cluster sizes to follow a power-law distribution.
Moreover, in their experiments, the authors set $\theta = \lambda/6$.
In this case, the objective function becomes:
\begin{align*}
	\arg \min_{l_1, ..., l_n} \sum_{c=1}^k \sum_{i \in l_c} \| \bm{x}_i - \bm{\mu}_c\| + \lambda \bigg(1 -\frac{ \ln k  }{6}\bigg) k,
\end{align*}
One can show that, when the number of data points exceeds $ e^6 \approx 403$, the trivial clustering result, namely every data point is a singleton cluster, will minimize this objective.
This can be seen by the fact that the trivial clustering result minimizes the $k$-means objective by simply being $0$ and that $k = \#$ of data points minimizes the regularization term.
In short, this objective is not appropriate for power-law clustering applications.
In the following experiment section, we will also compare our algorithm with their method empirically.


\section{Experiments}

We conclude with a brief set of experiments demonstrating the utility of our methods.
Namely, we will show that our approach enjoys benefits over the k-means algorithm on real power-law datasets in the vector setting and benefits over standard normalized cuts\footnote{Normalized cut image segmentation code: \\http://www.cis.upenn.edu/~jshi/software/.}  on synthetic and real data in the graph setting.
We also compare our method with the $pyp$-means~\cite{DBLP:journals/corr/FanZC13} and show that our method achieves better clustering results.
Throughout the experiments, we use normalized mutual information (NMI) between the algorithm's clusters and the ground-truth clusters for evaluation.

\begin{figure*}[t]
	\centering
      \begin{tabular}{ccc}
      		\includegraphics[scale=.20,natwidth=560,natheight=414]{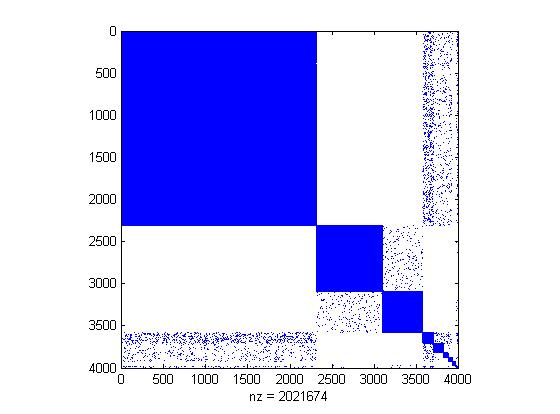} &
      		\includegraphics[scale=.20,natwidth=561,natheight=420]{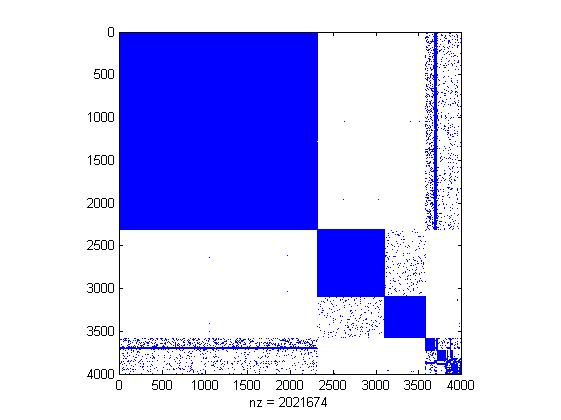} &
	      \includegraphics[scale=.20,natwidth=560,natheight=420]{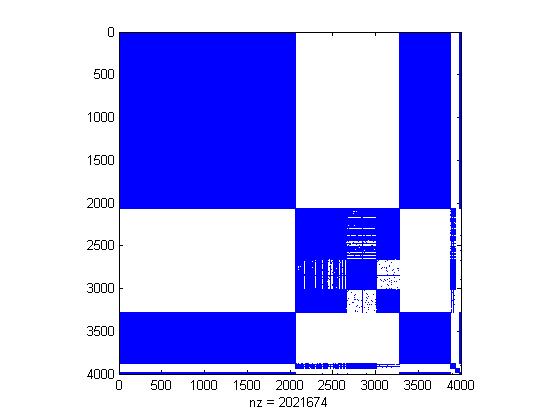}
      	\end{tabular}
      \caption{Results on a Pitman-Yor generated stochastic block model graph.  Left: Adjacency matrix of the graph, indexed by clusters.  Middle: Power-law normalized cut results; NMI is 0.866.  Right: Normalized Cuts result; NMI is 0.687.}
      \label{fig:syn_data_matrix}
\end{figure*}

\textbf{Synthetic power-law graph data.}
We begin with a synthetic power-law random graph dataset generated by the Pitman-Yor process applied to the stochastic block model.
Specifically, the Pitman-Yor CRP is first used to generate data cluster assignments and then a standard stochastic block model uses the assignments to generate a random graph.
We create a dataset with $4000$ nodes with $14$ disjoint clusters using the above process, with the corresponding adjacency matrix  shown in the left of Figure~\ref{fig:syn_data_matrix}.
The parameters  $\alpha$ and $\theta$ we use in the Pitman-Yor process model is $1$ and $0.2$ respectively.
In the stochastic block model, the stochastic block matrix is sampled from two Gaussian distributions: one being $\mathcal{N}(0.3, 0.001)$ for diagonal entries and the other being $\mathcal{N}(0.01, 0.001)$  for non-diagonal entries.
Our power-law normalized cut algorithm is then applied on this dataset with parameters validated on a separate validation dataset generated from the same process.
We compare with normalized cuts with its $k$ set to be the ground-truth. The results are shown in Figure \ref{fig:syn_data_matrix}; normalized cuts splits the big clusters while our algorithm nearly produces the ground-truth clusters.

\begin{table}[t]
\caption{NMI scores on a set of UCI power-law datasets.}
\label{tab:1}
\begin{center}
\begin{tabular}{cccc}
\toprule
 				&   \multicolumn{3}{c}{NMI}               	\\
\cmidrule(r){2-4}
 \bf Dataset				&	Ours 	&$k$-means	& $pyp$-means		 \\
\midrule
\emph{audiology}		&	0.621	&	0.518	&	0.417\\
\emph{ecoli}			&	0.700 	& 	0.545 	&	0.608\\
\emph{glass}			&	0.427 	&	0.315 	&	0.297\\
\emph{hypothyroid}		&	0.024 	&	0.009 	&  	0.077\\
\emph{page-blocks}		&	0.209 	&	0.123	&	0.088\\
\emph{flags}			&	0.275 	&	0.198 	&	0.178\\
\bottomrule
\end{tabular}
\end{center}
\end{table}

\begin{figure*}[t]
   \centering
      \begin{tabular}{ccc}
      \includegraphics[scale=.20]{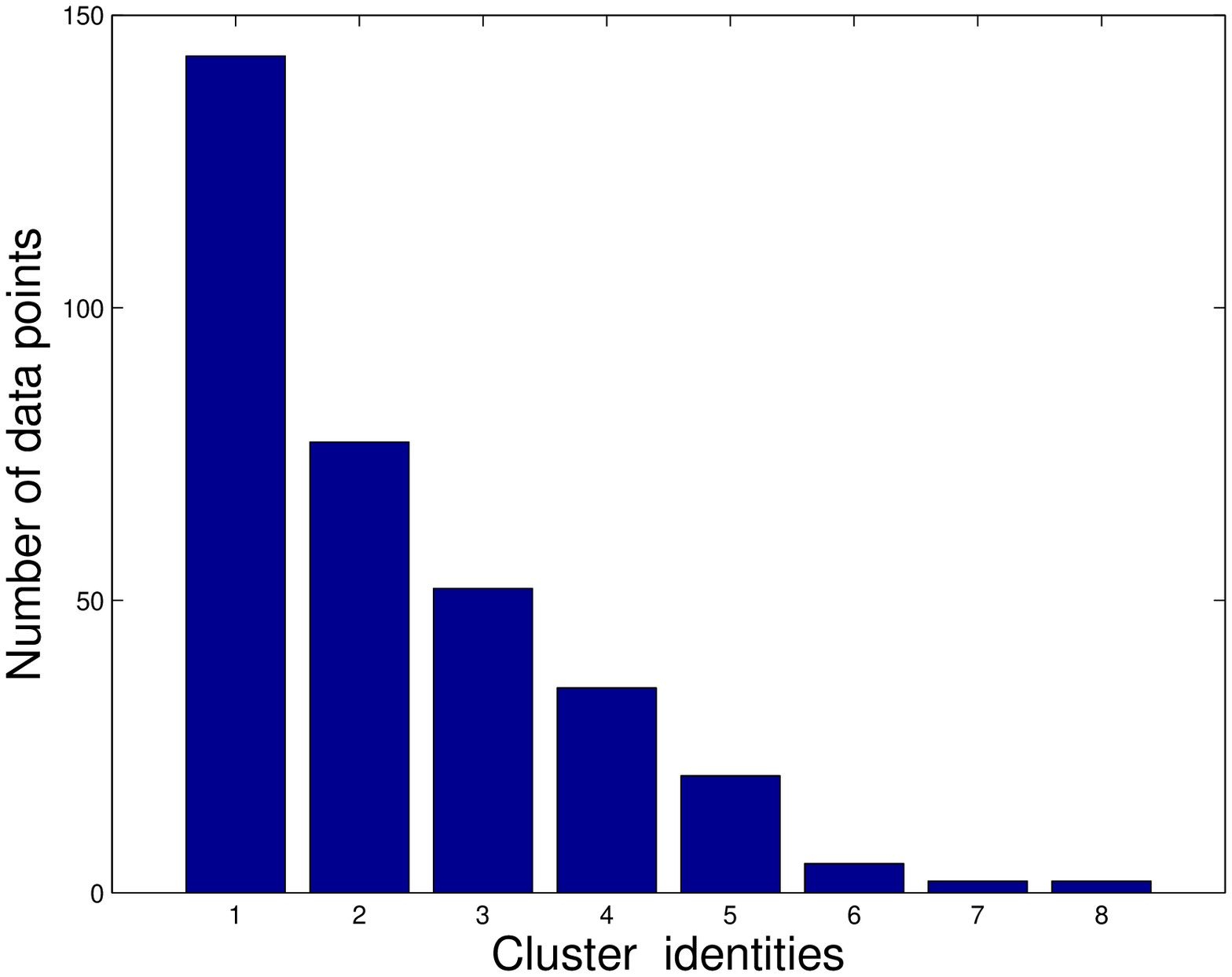} &
      \includegraphics[scale=.20]{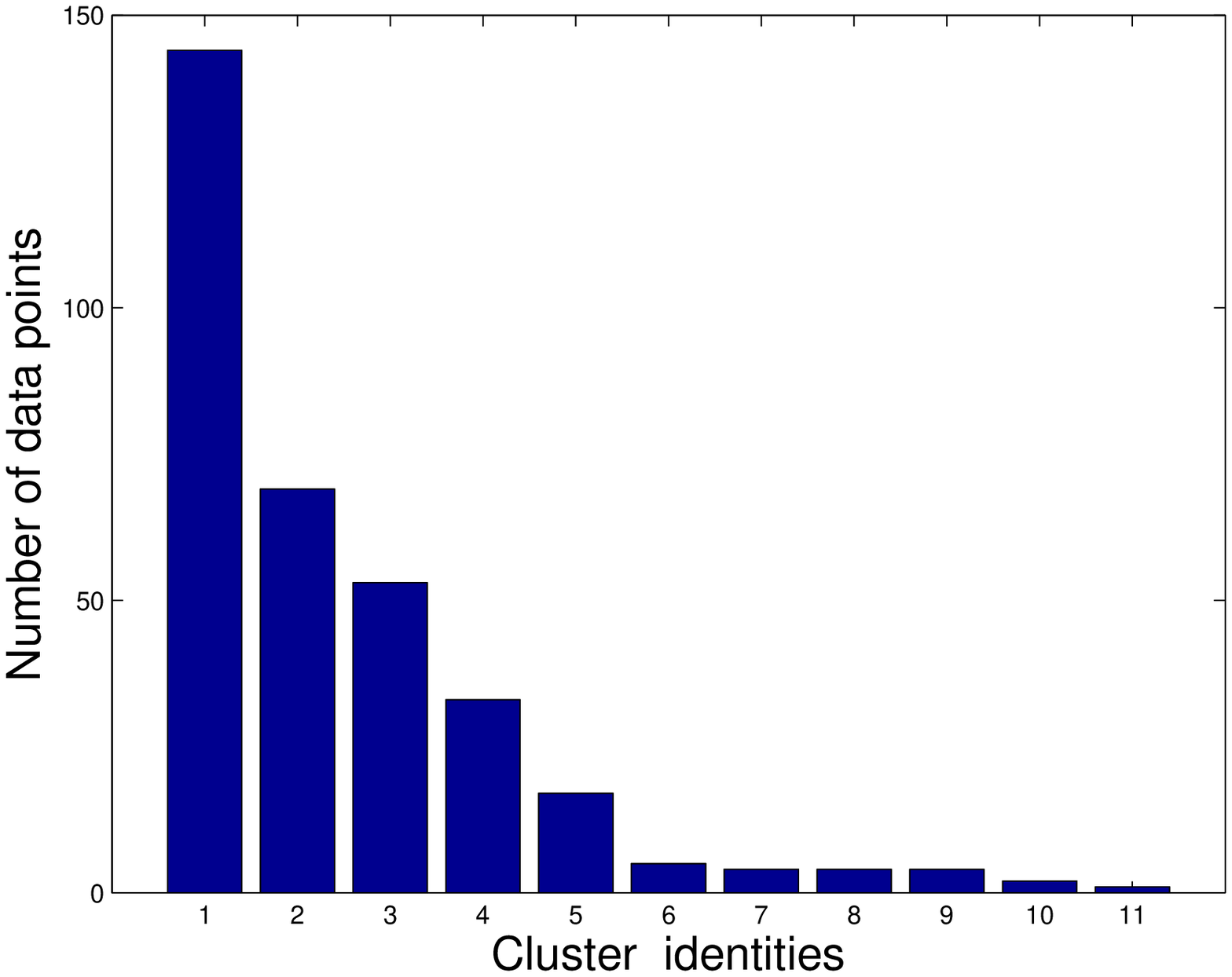} &
      \end{tabular}
      \begin{tabular}{ccc}
      \includegraphics[scale=.25]{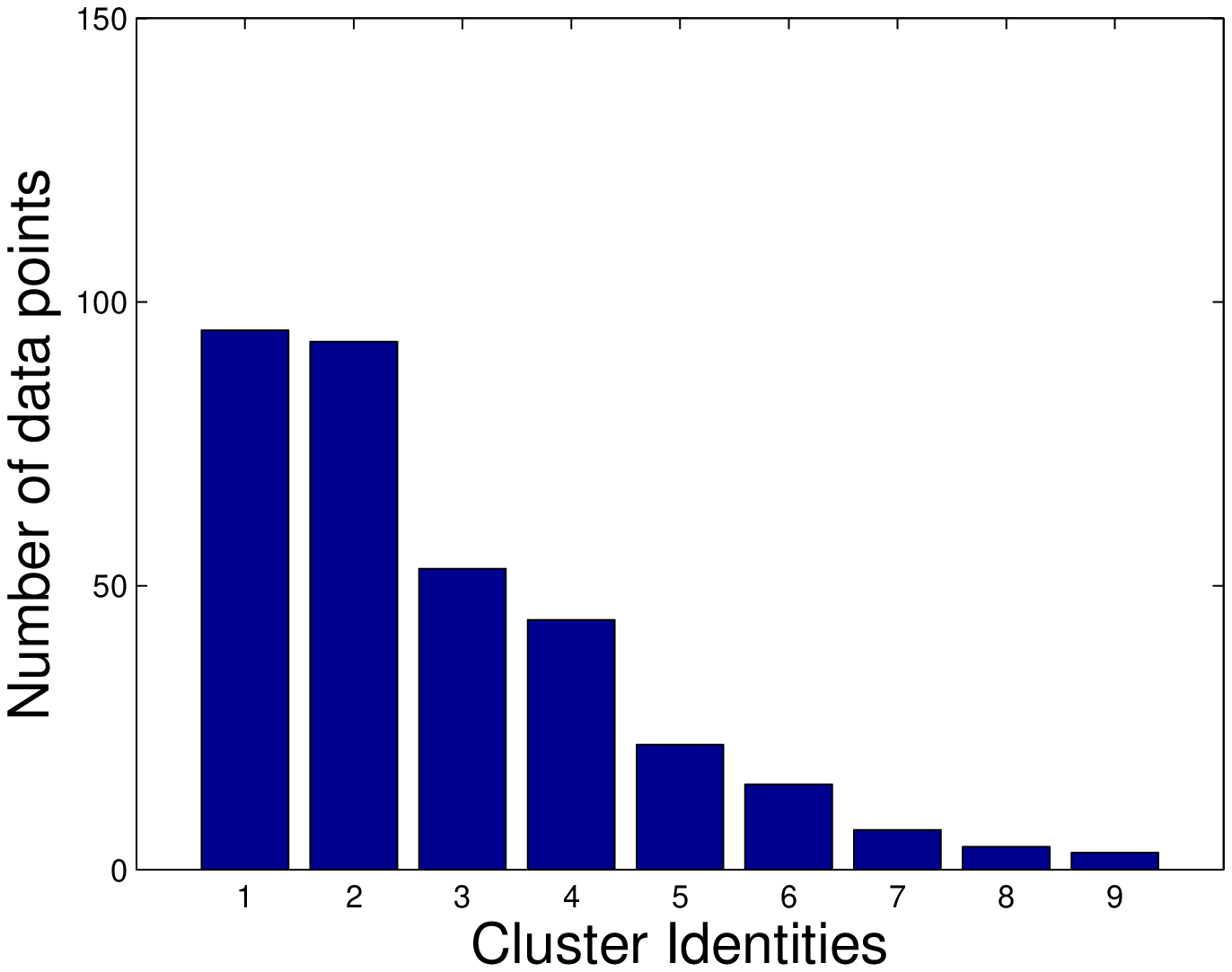}
      \includegraphics[scale=.264]{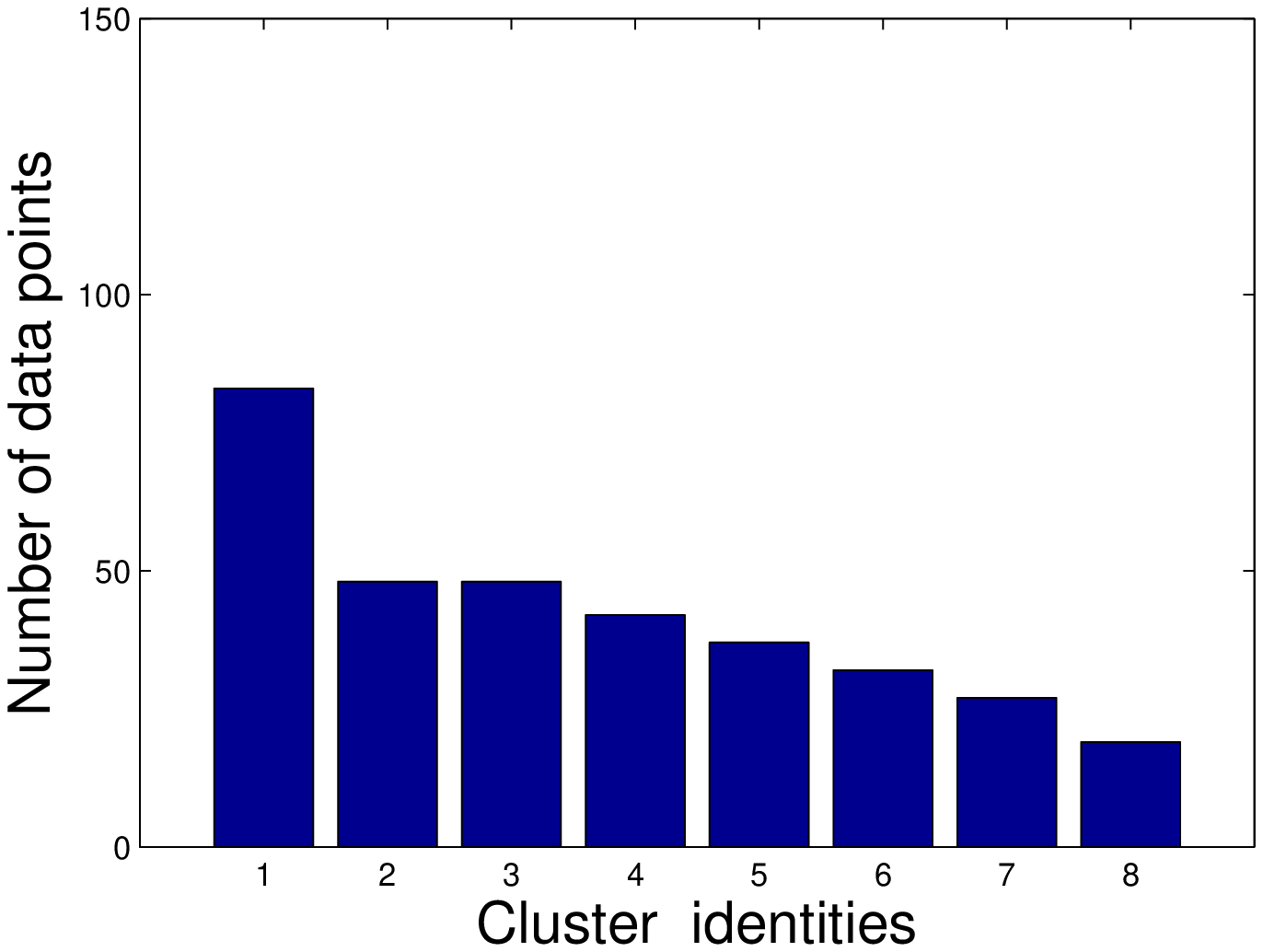}
      \end{tabular}
   \caption{Cluster size distributions on the \emph{ecoli} data set.  Upper left: Ground-truth. Upper right: Algorithm 1; NMI is 0.723. Lower left: pyp-means; NMI is 0.608. Lower right: k-means; NMI is 0.582.}
   \label{fig:power_law_uci}
\end{figure*}

\textbf{Real world power-law data sets.}\label{Sec:exp}
Next we consider comparing Algorithm \ref{algorithm} with k-means and $pyp$-means~\cite{DBLP:journals/corr/FanZC13} on real world benchmark data sets to demonstrate that our algorithm performs best on clustering vector data when cluster sizes are power-law distributed.
We selected $6$ UCI classification datasets whose class labels are power-law distributed (see Figure \ref{fig:power_law_uci}) and use class labels as the ground-truth for clusters.
Each dataset is then randomly split 30/70 for validation/clustering.
We normalize the datasets so that the values of all features lie in $[0,1]$.
On each validation set, we validate the parameters of Algorithm \ref{algorithm} (i.e. $\lambda, \alpha, \theta$) and the parameters of $pyp$-means only to yield cluster numbers close to the ground-truth (to make a fair comparison with k-means).
On each clustering set, we use the validated parameter settings for Algorithm \ref{algorithm} and $pyp$-means and  the ground-truth $k$ for k-means to perform the clustering.
The NMI are computed between the ground-truth and the computed clusters, and results are averaged over $10$ runs, as shown in Table \ref{tab:1}.
As we can see, Algorithm \ref{algorithm} performs better than k-means on all $6$ datasets in terms of NMI.
Also, it is better than $pyp$-means on all datasets except on the \textit{hypothyroid}.
Note that the $pyp$-means is better than $k$-means in $3$ datasets and worse than $k$-means in the other $3$.
Such high variance results on power-law datasets make us doubt that $pyp$-means is really able to achieve power-law clustering.
In Figure \ref{fig:power_law_uci}, we show the resulting clusterings on the \emph{ecoli} dataset given by Algorithm \ref{algorithm}, $pyp$-means and k-means, in which we use the whole dataset for clustering with validated parameters.
It is clear that k-means produces more uniform clusters and $pyp$-means also splits the largest cluster in the dataset.
\begin{table}[t]
\caption{NMI scores on graphs generated from UCI power-law datasets.}
\label{tab:3}
\begin{center}
\begin{tabular}{ccc}
\toprule
 				&   \multicolumn{2}{c}{NMI}  \\             	
\cmidrule(r){2-3}
 \bf Dataset	&	Ours 	&    Normalized cuts \\	
\midrule
\emph{ audiology}		&	0.662 &	0.561\\	
\emph{ ecoli}			&	0.702 & 	0.591\\ 	
\emph{ glass}		&	0.432 &	0.356\\  
\emph{ hypothyroid}	&	0.011 &	0.008\\  
\emph{ page-blocks}	&	0.222 &	0.126\\	
\emph{ flags}			&	0.357 &	0.200\\  
\bottomrule
\end{tabular}
\end{center}
\end{table}

\textbf{Real world power-law graph data sets.}\label{Sec:exp}
In this part we convert the UCI vector datasets used in the preceding experiment to form power-law graphs  and perform power-law normalized cuts on these graphs. We also run normalized cuts algorithm on these graphs to compare with our method.

To obtain the graphs, we first form the adjacency matrix by using a Gaussian similarity kernel on the vector data after normalizing them to $[0,1]$.
Then we use the adjacency matrix to form the kernel matrix and the weights as dsiccused in Section \ref{sec:powerlawncutalgorithm}. We randomly split data into validation/clustering with ratio of 30/70.
Parameters are selected on the validation set so that cluster numbers are close to the ground-truth.
The number of clusters in normalized cuts is set to the true number of clusters.
This is again for a fair comparison with normalized cuts.
Finally, we apply our power-law normalized cuts and normalized cuts on the clustering dataset.
NMI averaged over 10 runs are shown in Table \ref{tab:3}.

As we can see, our power-law normalized cuts is better than normalized cuts on all the graphs in terms of NMI.


\begin{figure}[t]
    \centering
        \includegraphics[width=54pt, height=36pt, natwidth=212, natheight=142]{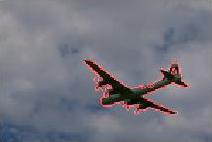}
   	    \includegraphics[width=54pt, height=36pt, natwidth=382, natheight=255]{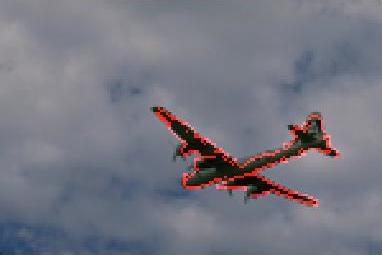}
        \includegraphics[width=54pt, height=36pt, natwidth=377,natheight=251]{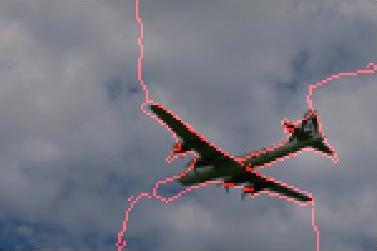}
      	\includegraphics[width=54pt, height=36pt, natwidth=346,natheight=233]{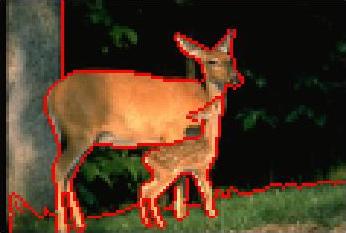}
		\includegraphics[width=54pt, height=36pt, natwidth=290,natheight=195]{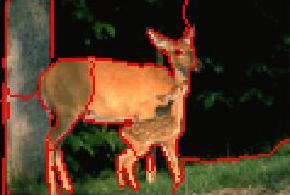}
		\includegraphics[width=54pt, height=36pt,natwidth=293,natheight=196]{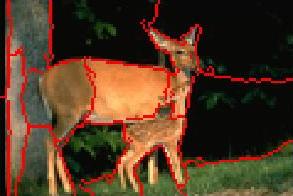}						
		\includegraphics[width=54pt, height=36pt,natwidth=255,natheight=173]{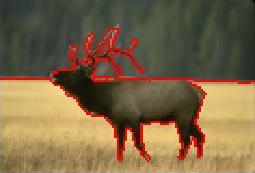}
     	\includegraphics[width=54pt, height=36pt,natwidth=343,natheight=231]{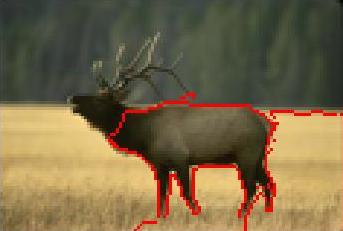}
      	\includegraphics[width=54pt, height=36pt,natwidth=261,natheight=147]{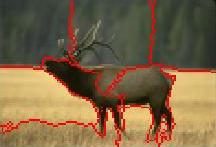}						
		\includegraphics[width=54pt, height=36pt,natwidth=288,natheight=193]{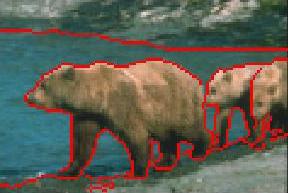}
     	\includegraphics[width=54pt, height=36pt,natwidth=352,natheight=283]{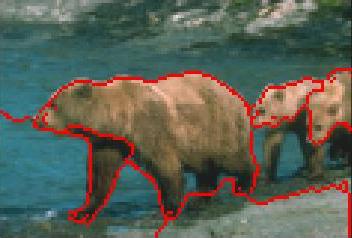}
      	\includegraphics[width=54pt, height=36pt,natwidth=252,natheight=171]{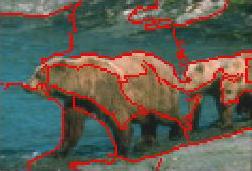}					
		\includegraphics[width=54pt, height=81pt,natwidth=149,natheight=226]{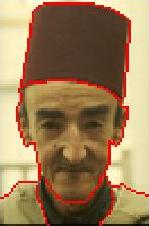}
     	\includegraphics[width=54pt, height=81pt,natwidth=146,natheight=221]{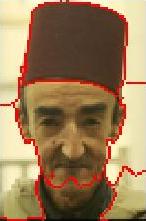}
      	\includegraphics[width=54pt, height=81pt,natwidth=172,natheight=261]{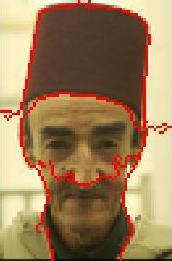}			
		\includegraphics[width=54pt, height=81pt,natwidth=245,natheight=364]{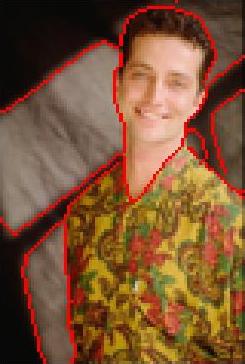}
     	\includegraphics[width=54pt, height=81pt,natwidth=154,natheight=229]{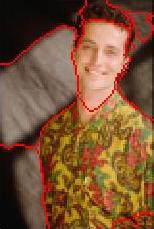}
      	\includegraphics[width=54pt, height=81pt,natwidth=209,natheight=311]{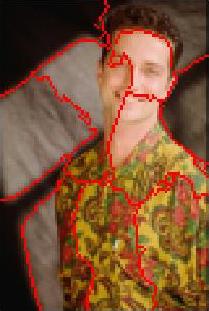}				
    \caption{Image segmentation results.  Left to right: ground-truth, our method, normalized cuts.}
    \label{fig:image_seg}
\end{figure}

\textbf{Image segmentation.}
Finally, we briefly demonstrate some qualitative results on image segmentation on the Berkeley segmentation data set~\cite{arbelaez2011contour}.
We adopt an approach that is similar to the approach in~\cite{shi2000normalized} to compute the affinity matrix.
Then we perform our power-law normalized cuts with the affinity matrix.
We compare standard normalized cuts with our proposed method on graphs generated from input images.
Figure~\ref{fig:image_seg} displays some example images; we see that normalized cuts tends to break up large segments more often than our approach.


\section{Conclusion}

We proposed a general framework of power-law graph cut algorithms that produce clusters whose sizes are power-law distributed, and also does not fix the number of clusters upfront.
The Pitman-Yor exchangeable partition probability function (EPPF) was incorporated into power-law graph cut objectives as a regularizer to promote power-law cluster size distributions.  A simple iterative algorithm was then proposed to locally optimize several objectives. Our proposed algorithm can be viewed as performing MAP inference on a particular Pitman-Yor mixture model. Finally, we conducted experiments on various data sets and showed the effectiveness of our algorithms against competing baselines.

\subsubsection*{References}
\bibliographystyle{unsrt}

\bibliography{reference}

\begin{thebibliography}{10}

\bibitem{shi2000normalized}
Jianbo Shi and Jitendra Malik.
\newblock Normalized cuts and image segmentation.
\newblock {\em Pattern Analysis and Machine Intelligence, IEEE Transactions
  on}, 22(8):888--905, 2000.

\bibitem{chan1994spectral}
Pak~K Chan, Martine~DF Schlag, and Jason~Y Zien.
\newblock Spectral k-way ratio-cut partitioning and clustering.
\newblock {\em Computer-Aided Design of Integrated Circuits and Systems, IEEE
  Transactions on}, 13(9):1088--1096, 1994.

\bibitem{dhillon2007weighted}
Inderjit~S Dhillon, Yuqiang Guan, and Brian Kulis.
\newblock Weighted graph cuts without eigenvectors a multilevel approach.
\newblock {\em IEEE Transactions on Pattern Analysis and Machine Intelligence},
  29(11):1944--1957, 2007.

\bibitem{greig1989exact}
DM~Greig, BT~Porteous, and Allan~H Seheult.
\newblock Exact maximum a posteriori estimation for binary images.
\newblock {\em Journal of the Royal Statistical Society. Series B
  (Methodological)}, pages 271--279, 1989.

\bibitem{sudderth2008shared}
Erik~B Sudderth and Michael~I Jordan.
\newblock Shared segmentation of natural scenes using dependent {P}itman-{Y}or
  processes.
\newblock In {\em NIPS}, pages 1585--1592, 2008.

\bibitem{clauset2009power}
Aaron Clauset, Cosma~Rohilla Shalizi, and Mark~EJ Newman.
\newblock Power-law distributions in empirical data.
\newblock {\em SIAM review}, 51(4):661--703, 2009.

\bibitem{hjort}
Nils Hjort, Chris Holmes, Peter Mueller, and Stephen Walker.
\newblock {\em Bayesian Nonparametrics: Principles and Practice}.
\newblock Cambridge University Press, Cambridge, UK, 2010.

\bibitem{pitman1997two}
Jim Pitman and Marc Yor.
\newblock The two-parameter {P}oisson-{D}irichlet distribution derived from a
  stable subordinator.
\newblock {\em The Annals of Probability}, pages 855--900, 1997.

\bibitem{pitman}
Jim Pitman.
\newblock {\em Combinatorial Stochastic Processes}.
\newblock Springer-Verlag, 2006.
\newblock Lectures from the Saint-Flour Summer School on Probability Theory.

\bibitem{kulis2012revisiting}
Brian Kulis and Michael~I Jordan.
\newblock Revisiting k-means: New algorithms via {B}ayesian nonparametrics.
\newblock In {\em Proceedings of the 29th International Conference on Machine
  Learning (ICML-12)}, pages 513--520, 2012.

\bibitem{broderick2012mad}
Tamara Broderick, Brian Kulis, and Michael~I Jordan.
\newblock {MAD}-{b}ayes: {MAP}-based asymptotic derivations from {B}ayes.
\newblock In {\em Proceedings of the 30th International Conference on Machine
  Learning (ICML-13)}, 2013.

\bibitem{DBLP:journals/corr/FanZC13}
Xuhui Fan, Yiling Zeng, and Longbing Cao.
\newblock Non-parametric power-law data clustering.
\newblock {\em CoRR}, abs/1306.3003, 2013.

\bibitem{kernighan1970efficient}
Brian~W Kernighan and Shen Lin.
\newblock An efficient heuristic procedure for partitioning graphs.
\newblock {\em Bell system technical journal}, 49(2):291--307, 1970.

\bibitem{yu2003multiclass}
Stella~X Yu and Jianbo Shi.
\newblock Multiclass spectral clustering.
\newblock In {\em Computer Vision, 2003. Proceedings. Ninth IEEE International
  Conference on}, pages 313--319. IEEE, 2003.

\bibitem{arbelaez2011contour}
Pablo Arbelaez, Michael Maire, Charless Fowlkes, and Jitendra Malik.
\newblock Contour detection and hierarchical image segmentation.
\newblock {\em Pattern Analysis and Machine Intelligence, IEEE Transactions
  on}, 33(5):898--916, 2011.

\end{thebibliography}
\end{document}